\documentclass[final]{cvpr}

\usepackage{times}
\usepackage{epsfig}
\usepackage{graphicx}
\usepackage{amsmath}
\usepackage{amssymb}
\usepackage{color}
\usepackage{subcaption}
\usepackage{mathtools}
\usepackage{paralist} % compactenum
\usepackage{enumitem}
\usepackage{colortbl}
\usepackage{nicefrac}
\usepackage{multirow}
\usepackage[table,dvipsnames]{xcolor}
\usepackage{booktabs}
\usepackage{makecell}
\usepackage{gensymb}
\usepackage{microtype}
\usepackage{cancel}
\usepackage{wasysym}

\definecolor{yellow}{rgb}{1,1, 0.7}
\definecolor{orange}{rgb}{1, 0.8, 0.6}
\definecolor{red}{rgb}{1, 0.6, 0.6}
\definecolor{tangerine}{rgb}{0.95, 0.52, 0.}

\definecolor{darkyellow}{rgb}{0.8, 0.8, 0.5}
\definecolor{darkred}{rgb}{0.7, 0.3, 0.3}
\definecolor{darkgreen}{rgb}{0.3, 0.7, 0.3}
\definecolor{blue}{rgb}{0, 0, 1.0}
\definecolor{green}{rgb}{0, 1.0, 0}
\definecolor{pink}{rgb}{1, 0.4, 0.7}

\newcommand{\best}[1]{\cellcolor{yellow} #1}

\let\originalleft\left
\let\originalright\right
\renewcommand{\left}{\mathopen{}\mathclose\bgroup\originalleft}
\renewcommand{\right}{\aftergroup\egroup\originalright}

\newcommand{\norm}[1]{\left\lVert#1\right\rVert}

\newcommand{\expo}[1]{\exp\left(#1\right)}

\newcommand{\absrp}{\sigma}

\DeclareMathSymbol{\shortminus}{\mathbin}{AMSa}{"39}
\newcommand{\vismult}{\lambda}

\usepackage[pagebackref=true,breaklinks=true,colorlinks,bookmarks=false]{hyperref}

\usepackage[font=small]{caption}

\def\cvprPaperID{2454} % *** Enter the CVPR Paper ID here
\def\confYear{CVPR 2021}

\begin{document}

%%%%%%%%% TITLE
\title{NeRV: Neural Reflectance and Visibility Fields for Relighting and View Synthesis}

\author{
\and
Pratul P. Srinivasan\\
\small{Google Research}
\and
Boyang Deng\\
\small{Google Research}
\and
Xiuming Zhang\\
\small{MIT}
\and
Matthew Tancik\\
\small{UC Berkeley}
\and
\and
Ben Mildenhall\\
\small{UC Berkeley}
\and
Jonathan T. Barron\\
\small{Google Research}
}

\maketitle

%%%%%%%%% ABSTRACT
\begin{abstract}

We present a method that takes as input a set of images of a scene illuminated by unconstrained known lighting, and produces as output a 3D representation that can be rendered from novel viewpoints under arbitrary lighting conditions. Our method represents the scene as a continuous volumetric function parameterized as MLPs whose inputs are a 3D location and whose outputs are the following scene properties at that input location: volume density, surface normal, material parameters, distance to the first surface intersection in any direction, and visibility of the external environment in any direction. Together, these allow us to render novel views of the object under arbitrary lighting, including indirect illumination effects. The predicted visibility and surface intersection fields are critical to our model's ability to simulate direct and indirect illumination during training, because the brute-force techniques used by prior work are intractable for lighting conditions outside of controlled setups with a single light. Our method outperforms alternative approaches for recovering relightable 3D scene representations, and performs well in complex lighting settings that have posed a significant challenge to  prior work.

\end{abstract}

%%%%%%%%% BODY TEXT
\section{Introduction}

\begin{figure}[t]
\begin{center}
    \includegraphics[width=1\linewidth]{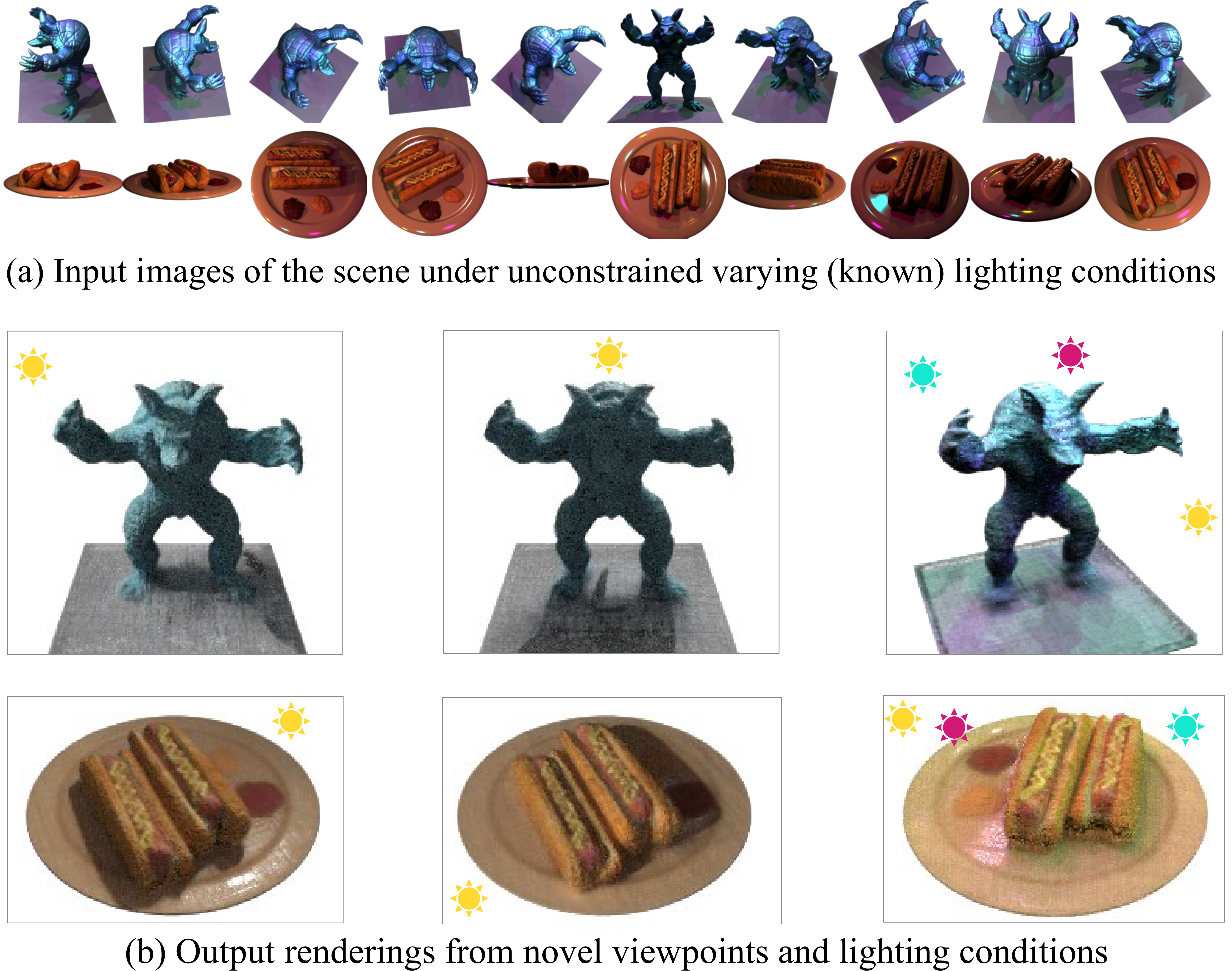}
\end{center}
   \vspace{-0.2in}
   \caption{We optimize a Neural Reflectance and Visibility Field (NeRV) 3D representation from a set of input images of a scene illuminated by known but unconstrained lighting. Our NeRV representation can be rendered from novel views under arbitrary lighting conditions not seen during training. Here, we visualize example input data and renderings for two scenes. The first two output rendered images for each scene are from the same viewpoint, each illuminated by a point light at a different location, and the last image is from a different viewpoint under a random colored illumination.
   }
   \label{fig:teaser}
\end{figure}

A central problem in computer vision is that of inferring the physical geometry and material properties that together explain observed images. In addition to its importance for recognition and robotics, a solution to this open problem would have significant value for computer graphics --- the ability to create realistic 3D models from standard photos could democratize 3D content creation and allow anyone to use real-world objects in photography, filmmaking, and game development.
In this paper, we work towards this goal and present an approach for estimating a volumetric 3D representation from images of a scene under arbitrary known lighting conditions, such that high-quality novel images can be rendered from arbitrary unseen viewpoints and under novel unobserved lighting conditions, as shown in Figure~\ref{fig:teaser}.

The vision and graphics research communities have recently made substantial progress towards the novel view synthesis portion of this goal. 
The Neural Radiance Fields (NeRF)~\cite{mildenhall2020nerf} approach has shown that it is possible to synthesize photorealistic images of scenes by training a simple neural network to map 3D locations in the scene to a continuous field of volume density and color.
Volume rendering is trivially differentiable, so the parameters of a NeRF can be optimized for a single scene by using gradient descent to minimize the difference between renderings of the NeRF and a set of observed images. Though NeRF produces compelling results for view synthesis, it does not provide a solution for relighting. 
This is because NeRF models just the amount of outgoing light from a location --- the fact that this outgoing light is the result of interactions between incoming light and the material properties of an  underlying surface is ignored.

\begin{figure}[t]
\begin{center}
    \includegraphics[width=1\linewidth]{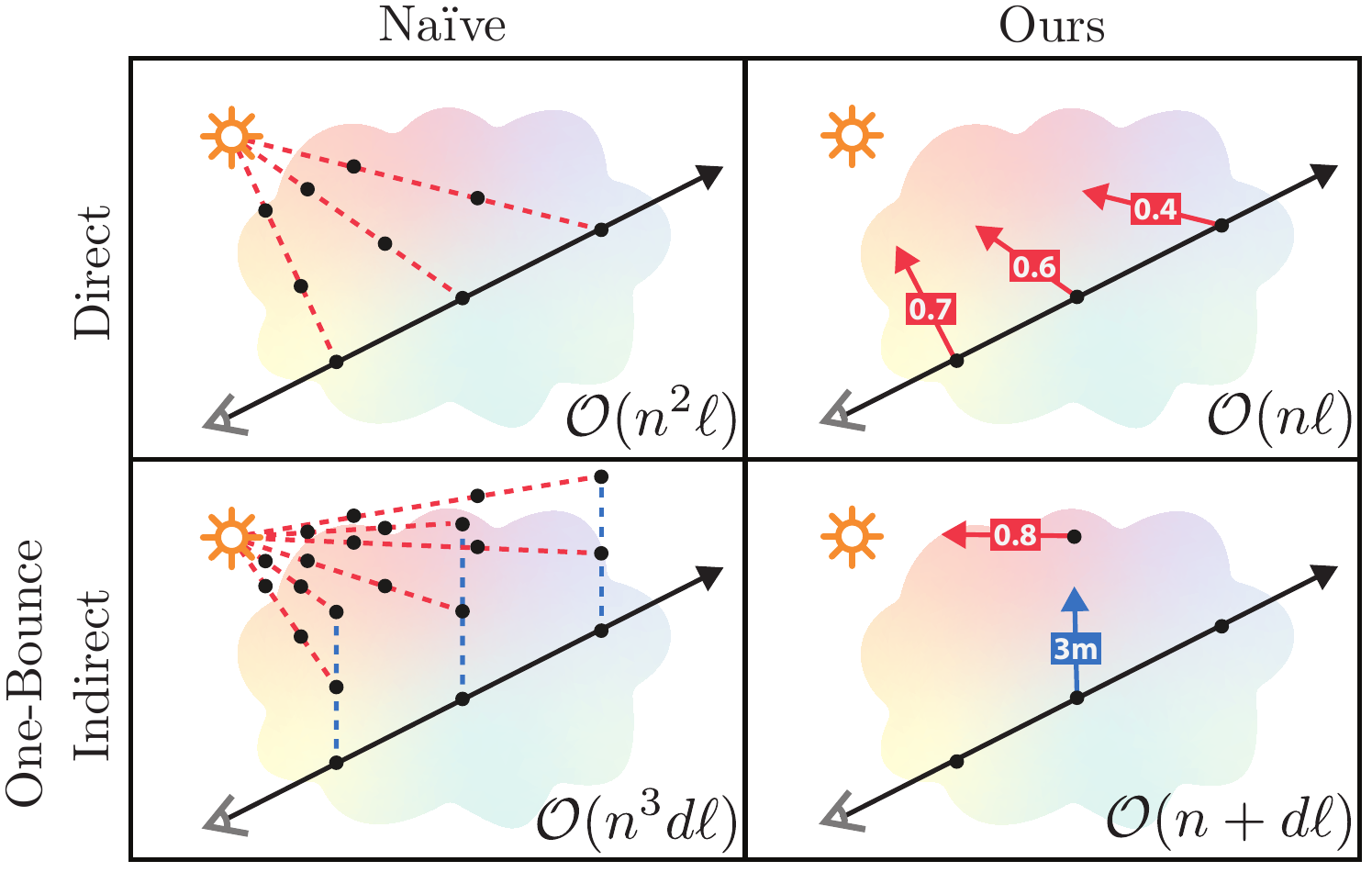}
\end{center}
\vspace{-0.15in}
   \caption{
   We visualize how ``Neural Visibility Fields'' reduce the computational burden of volume rendering a camera ray with direct (top) and one-bounce indirect (bottom) illumination compared to na\"ive raymarching, alongside each solution's computational complexity ($n$ is the number of samples along each ray, $\ell$ is the number of light sources, and $d$ is the number of sampled indirect illumination directions).
   Black dots represent evaluating a shape MLP for volume density at a position, red arrows represent evaluating the visibility MLP at a position along a direction, and the blue arrow represents evaluating the visibility MLP for the expected termination depth of a ray at a position along a direction (output visibility multipliers and termination depths from the visibility MLP are displayed as text).
   Brute-force light transport simulation through NeRF's volumetric representation with na\"ive raymarching (left) is intractable.
   By approximating visibility with a neural visibility field (right) that is optimized alongside the shape MLP, we are able to make optimization with complex illumination tractable.
   }
   \label{fig:problem}
\end{figure}

At first glance, extending NeRF to enable relighting appears to require only changing the image formation model: instead of modeling scenes as fields of density and view-dependent color, we can model surface normals and material properties (\eg the parameters of a bi-directional reflectance distribution function (BRDF)) and simulate the transport of the scene's light sources (which we assume are known) according to the rules of physically based rendering~\cite{PBR}. 
However, simulating the attenuation and reflection of light by particles is fundamentally challenging in NeRF's neural volumetric representation because content can exist \emph{anywhere} within the scene, and determining the density at any location requires querying a neural network. Consider the na\"ive procedure for computing the radiance along a single camera ray due to direct illumination, as illustrated in Figure~\ref{fig:problem}: First, we query NeRF's multi-layer perceptron (MLP) for the volume density at samples along the camera ray to determine the amount of light reflected by particles at each location that reaches the camera. For each location along the camera ray, we then query the MLP for the volume density at densely-sampled points between the location and every light source to estimate the attenuation of light before it reaches that location. This procedure quickly becomes prohibitively expensive if we want to model environment light sources or global illumination, in which case scene points may be illuminated from all directions. Prior methods for estimating relightable volumetric representations from images have not overcome this challenge, and can only simulate direct illumination from a single point light source when training.

The problem of efficiently computing visibility is well explored in the graphics literature. In standard raytracing graphics pipelines, where the scene geometry is fixed and known ahead of time, a common solution is to precompute a data structure that can be efficiently queried to obtain 
the visibility between pairs of scene points, or between scene points and light sources.
This can be accomplished with approaches including octrees~\cite{samet1989implementing}, distance transforms~\cite{cohen1994proximity}, or bounding volume hierarchies~\cite{PBR}.
But these existing approaches do not provide a solution to our task --- our geometry is unknown, and our model's estimate of geometry changes constantly as it is optimized.
Though conventional data structures could perhaps be used to accelerate rendering \emph{after} optimization is complete, we need to efficiently query the visibility between points \emph{during} optimization, and existing solutions are prohibitively expensive to rebuild after each training iteration (of which there may be millions).

In this work, we present a method to train a NeRF-like model that can simulate realistic environment lighting and global illumination. Our key insight is to train an MLP to act as a lookup table into a \emph{visibility field} during rendering. Instead of estimating light or surface visibility at a given 3D position along a given direction by densely evaluating an MLP for the volume density along the corresponding ray (which would be prohibitively expensive), we simply query this visibility MLP to estimate visibility and expected termination depth in any direction (see Figure~\ref{fig:problem}). This visibility MLP is optimized alongside the MLP that represents volume density, and is supervised to be consistent with the volume density samples observed during optimization. Using this neural approximation of the true visibility field significantly eases the computational burden of estimating volume rendering integrals while training.
Our resulting system, which we call ``NeRV'' (``Neural Reflectance and Visibility Fields'') enables the recovery of a NeRF-like model that supports relighting in addition to view synthesis.
While previous solutions for relightable NeRFs~\cite{bi2020neural} were limited to controlled settings which required input images to be illuminated by a single point light, NeRV supports training with arbitrary environment lighting and ``one-bounce'' indirect illumination.

\section{Related Work}

Neural Radiance Fields~\cite{mildenhall2020nerf} can be thought of as a modern neural reformulation of the classic problem of scene reconstruction: given multiple images of a scene, inferring the underlying geometry and appearance that best explains those images. While classic approaches have largely relied on discrete representations such as textured meshes~\cite{Hartley2003, PhotoTourism} and voxel grids~\cite{seitz99}, NeRF has demonstrated that a continuous volumetric function, parameterized as an MLP, is able to represent complex scenes and render photorealistic novel views. 
NeRF works well for view synthesis, but it does not enable relighting because it has no mechanism to disentangle the outgoing radiance of a surface into an incoming radiance and an underlying surface material. 

This problem of attributing what aspects of an image are due to material, lighting, or geometric variation is commonly referred to as ``intrinsic image estimation''~\cite{Barrow1978, Land71lightnessand} or ``inverse rendering''~\cite{ramamoorthi01,sato97}, and is a classic problem in computer vision and graphics. These classical approaches have derived insightful observations about separating a single image into shading and reflectance components~\cite{Horn1974DeterminingLF}, inferring surface normals from the appearance of an image's shading~\cite{Horn1970}, or jointly inferring shape, illumination, and reflectance from a single image~\cite{BarronTPAMI2015}, but they are not designed to recover full 3D models that can be used for graphics applications.

The difficulty of this problem (a consequence of its underconstrained nature) is typically addressed using one of the following strategies: 1) learning priors on shape, illumination, and reflectance, 2) assuming known geometry, or 3) using multiple input images of the scene under different lighting conditions. Most recent single-image inverse rendering methods~\cite{li2020inverse,li2018learning,sangsingle,sengupta19,wei2020object,yu2019inverserendernet} belong to the first category, and use large datasets of images with labeled geometry and materials to train convolutional neural networks to predict these properties. Most prior works in inverse rendering that recover full 3D models for graphics applications~\cite{weinmann15} fall under the second category, and use 3D geometry obtained from active scanning~\cite{park20chips,schmitt2020joint,zhang2020neural}, proxy models~\cite{chen2020neural,dong2014appearance,gao20}, silhouette masks~\cite{oxholm2014multiview,xia2016recovering}, or multiview stereo~\cite{nam2018practical} as a starting point before recovering reflectance and refined geometry.

Our method belongs to the third category; we only require posed input images of a scene under different (known) lighting conditions. The most related prior works are Deep Reflectance Volumes~\cite{bi2020deep}, which estimates voxel geometry and BRDF parameters, and the follow-up work Neural Reflectance Fields~\cite{bi2020neural}, which replaces Deep Reflectance Volume's voxel grid with a continuous volume represented by an MLP. Our work extends Neural Reflectance Fields (which requires scenes to only be illuminated by a single point light at a time due to their brute-force visibility computation strategy visualized in Figure~\ref{fig:problem}, and only models direct illumination) to work for arbitrary lighting and global illumination.

We also take inspiration from recent works that replace discrete voxel and mesh geometry representations with MLPs that approximate a continuous 3D function by mapping from an input 3D location to scene properties at that location. This strategy has been explored for the tasks of representing  shapes~\cite{neuralarticulated,learningshape,occupancynet,park2019deepsdf,sitzmann2020implicit,tancik2020fourfeat} and scenes under fixed lighting for view synthesis~\cite{nsvf,mildenhall2020nerf,diffvolumetric,srn,yariv20}. One technique that has been used for relighting these neural representations is to condition the MLP's output appearance on a latent code that encodes a per-image lighting, as in NeRF in the Wild~\cite{martin2020nerf} (as well as previously with discretized scene representations~\cite{li2020crowdsampling,meshry19}). Although this strategy can effectively explain the appearance variation of training images, it cannot be used to render the same scene under new lighting conditions not observed during training (Figure~\ref{fig:latent_comparison}) because it does not utilize the physics of light transport.

Our method is inspired by a long line of work in graphics that explores precomputation~\cite{ritschel07,sloan2002precomputed} and approximation~\cite{bun04,green07,ritschel09,ritschel08} strategies to efficiently compute global illumination in physically-based rendering. Our ``Neural Visibility Fields'' can be thought of as a neural analogue to visibility precomputation techniques, and is specifically designed for use in our neural inverse rendering setting where geometry is dynamically changing during optimization.

\section{Method}

\begin{figure*}[!t]
\begin{center}
    \includegraphics[width=1\linewidth]{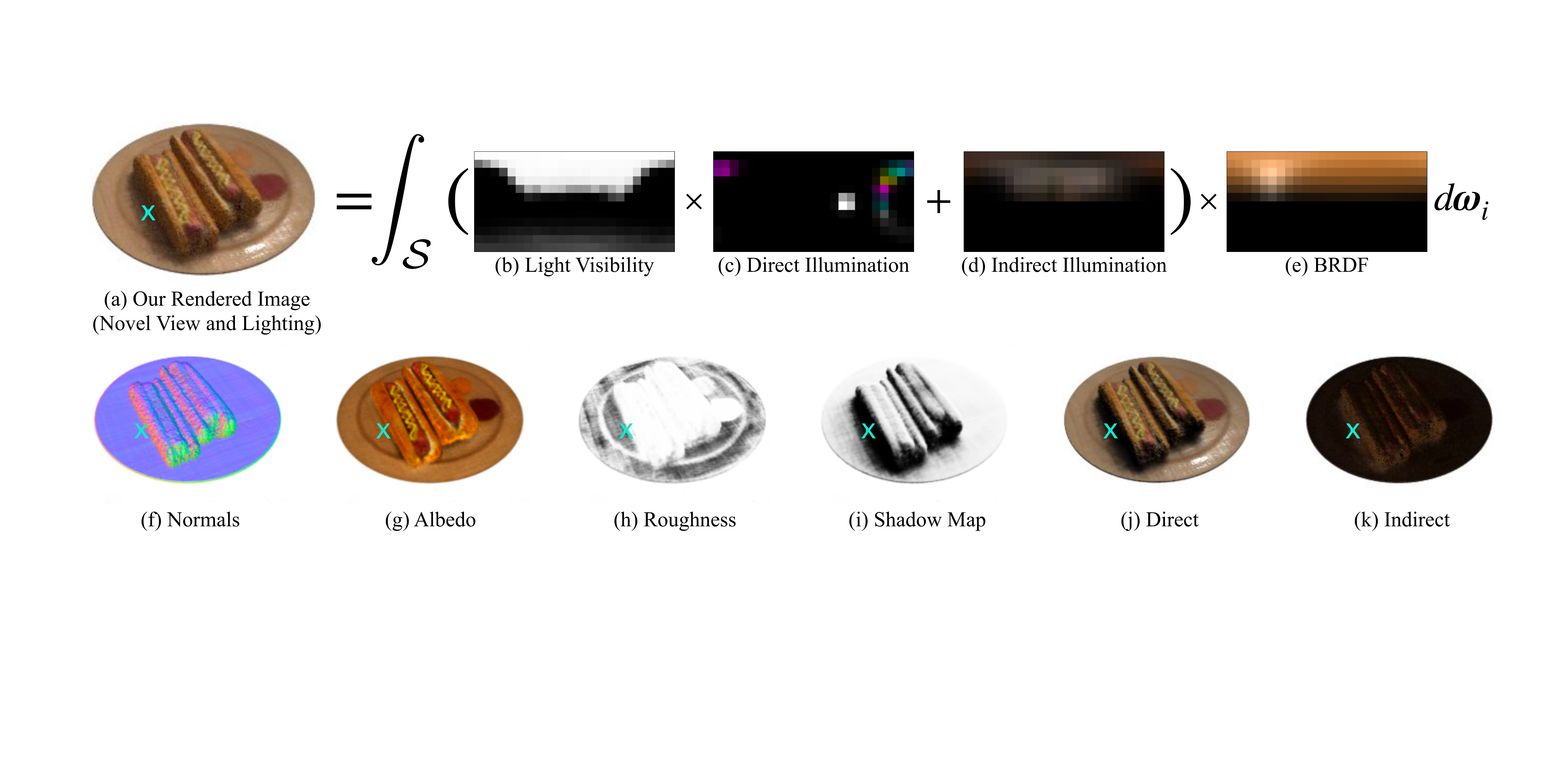}
\end{center}
\vspace{-0.2in}
   \caption{(a) Given any continuous 3D location as input, such as the point at the cyan ``x'', NeRV outputs the volume density as well as: 
   (b) The visibility to a spherical environment map surrounding the scene, which is multiplied by (c) the direct illumination at that point and added to (d) the estimated indirect illumination at that point to determine the full incident illumination. This is then multiplied by (e) the predicted BRDF, and integrated over all incoming directions to determine the outgoing radiance at that point. In the bottom row, we visualize these outputs for the full rendered image: (f) Surface normals, and BRDF parameters for (g) diffuse albedo and (h) specular roughness. We can use the predicted visibilities to compute the fraction of the total illumination that is actually incident at any location, visualized as (i) a shadow map. We also show the same rendered viewpoint if it were lit by only (j) direct and (k) indirect illumination.}
   \label{fig:hotdog}
\end{figure*}

We extend NeRF to include the simulation of light transport, which allows NeRFs to be rendered under arbitrary novel illumination conditions. Instead of modeling a scene as a continuous 3D field of particles that absorb and \emph{emit} light as in NeRF, we represent a scene as a 3D field of \emph{oriented} particles that absorb and \emph{reflect} the light emitted by external light sources (Section~\ref{sec:relightable}). Na\"ively simulating light transport through this model is inefficient and unable to scale to simulate realistic lighting conditions or global illumination. We remedy this by introducing a neural visibility field representation (optimized alongside NeRF's volumetric representation) that allows us to efficiently query the point-to-light and point-to-point visibilities needed to simulate light transport (Section~\ref{sec:visibility}). The resulting Neural Reflectance and Visibility Field (NeRV) is visualized in Figure~\ref{fig:hotdog}.

\subsection{NeRF Overview}
\label{sec:nerfoverview}

NeRF represents a scene as a continuous function, parameterized by a ``radiance'' MLP whose input is a 3D position and viewing direction, and whose output is the volume density $\absrp$ and radiance $L_e$ (RGB color) emitted by particles at that location along that viewing direction. NeRF uses standard emission-absorption volume rendering~\cite{kajiya84} to compute the observed radiance $L(\mathbf{c},\boldsymbol{\omega}_o)$ (the rendered pixel color) at camera location $\mathbf{c}$ along direction $\boldsymbol{\omega}_o$ as the integral of the product of three quantities at any point $\mathbf{x}(t)=\mathbf{c} - t\boldsymbol{\omega}_o$ along the ray: the visibility $V(\mathbf{x}(t),\mathbf{c})$, which indicates the fraction of emitted light from position $\mathbf{x}(t)$ that reaches the camera at $\mathbf{c}$, the density $\absrp(\mathbf{x}(t))$, and the emitted radiance $L_e(\mathbf{x}(t),\boldsymbol{\omega}_o)$ along the viewing direction $\boldsymbol{\omega}_o$:
\begin{gather}
\resizebox{0.9\linewidth}{!}{$\displaystyle%
L(\mathbf{c},\boldsymbol{\omega}_o) = \int_{0}^{\infty}\!\!V(\mathbf{x}(t),\mathbf{c})\absrp(\mathbf{x}(t))L_e(\mathbf{x}(t),\boldsymbol{\omega}_o)\,dt\,, \label{eqn:nerf_render1}
$}\\
V(\mathbf{x}(t),\mathbf{c}) = \expo{-\int_{0}^{t}\absrp(\mathbf{x}(s))\,ds}\,. \label{eqn:nerf_render2}
\end{gather}
A NeRF is recovered from observed input images of a scene by sampling a batch of observed pixels, sampling the corresponding camera rays of those pixels at stratified random points to approximate the above integral using numerical quadrature~\cite{max95}, and optimizing the weights of the radiance MLP via gradient descent to minimize the error between the estimated and observed pixel colors.

\subsection{Neural Reflectance Fields}
\label{sec:relightable}

A NeRF representation does not separate the effect of incident light from the material properties of surfaces. This means that NeRF is only able to render views of a scene under the fixed lighting conditions presented in the input images --- a NeRF cannot be relit. Modifying NeRF to enable relighting is straightforward, as initially demonstrated by the Neural Reflectance Fields work of Bi \etal~\cite{bi2020neural}. Instead of representing a scene as a field of particles that emit light, it is represented as a field of particles that reflect incoming light. With this, given an arbitrary lighting condition, we can simulate the transport of light through the volume as it is reflected by particles until it reaches the camera with a standard volume rendering integral~\cite{kajiya84}:
\begin{align}
L(\mathbf{c},\boldsymbol{\omega}_o) &= \int_{0}^{\infty}\!\!V(\mathbf{x}(t),\mathbf{c})\absrp(\mathbf{x}(t))L_r(\mathbf{x}(t),\boldsymbol{\omega}_o)\,dt\,, \label{eqn:reflect_render1} \\ 
L_r(\mathbf{x},\boldsymbol{\omega}_o) &= \int_{\mathcal{S}}L_i(\mathbf{x},\boldsymbol{\omega}_i)R(\mathbf{x},\boldsymbol{\omega}_i,\boldsymbol{\omega}_o)\,d\boldsymbol{\omega}_i\,, \label{eqn:reflect_render2}
\end{align}
where the view-dependent emission term $L_e(\mathbf{x},\boldsymbol{\omega_o})$ in Equation~\ref{eqn:nerf_render1} is replaced with an integral over the sphere $\mathcal{S}$ of incoming directions, of the product of the incoming radiance $L_i$ from any direction and a reflectance function $R$ (often called a phase function in volume rendering) which describes how much light arriving from direction $\boldsymbol{\omega}_i$ is reflected towards direction $\boldsymbol{\omega}_o$. We follow Bi \etal and use the standard microfacet BRDF described by Walter \etal~\cite{walter2007microfacet} as the reflectance function, so $R$ at any 3D location is parameterized by a diffuse RGB albedo, a scalar specular roughness, and a surface normal. We replace NeRF's radiance MLP with two MLPs: a ``shape'' MLP that outputs volume density $\absrp$ and a ``reflectance'' MLP that outputs BRDF parameters (3D diffuse albedo $\mathbf{a}$ and 1D roughness $\gamma$) for any input 3D point: $\text{MLP}_{\theta}:\mathbf{x} \to \sigma,\, \text{MLP}_{\psi}:\mathbf{x} \to (\mathbf{a}, \gamma)$.
Instead of parameterizing the 3D surface normal $\mathbf{n}$ as a normalized output of the shape MLP, as in Bi \etal~\cite{bi2020neural}, we compute $\mathbf{n}$ as the negative normalized gradient vector of the shape MLP's output $\absrp$ with respect to $\mathbf{x}$, computed using automatic differentiation. We further discuss this choice in Section~\ref{sec:ablations}. 

\begin{figure}[t]
\begin{center}
    \includegraphics[width=1\linewidth]{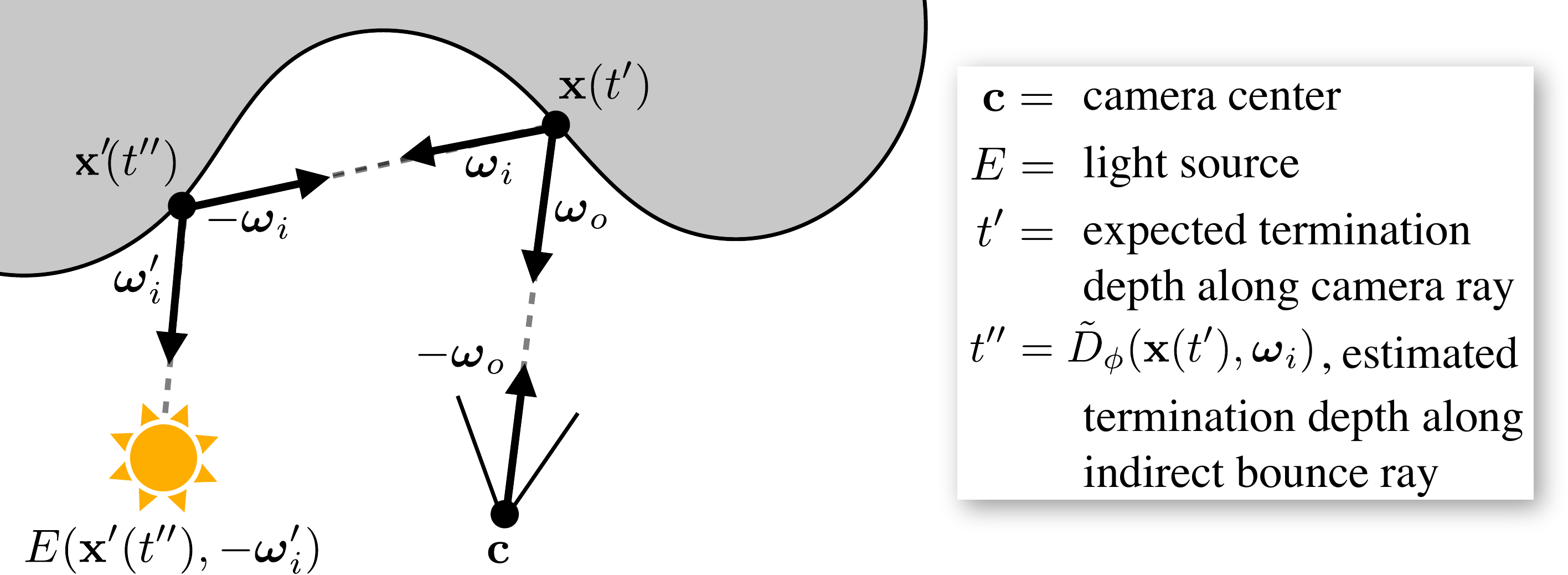}
\end{center}
\vspace{-0.25in}
   \caption{
   The geometry of an indirect illumination path from camera to light source, and a visualization of our notation. 
   }
   \label{fig:bean}
\end{figure}

\subsection{Light Transport via Neural Visibility Fields}
\label{sec:visibility}

Although modifying NeRF to enable relighting is straightforward, estimating the volume rendering integral for general lighting scenarios is computationally challenging with a continuous volumetric representation such as NeRF. Figure~\ref{fig:problem} visualizes the scaling properties that make simulating volumetric light transport particularly difficult. Even if we only consider direct illumination from light sources to a scene point, a brute-force solution is already challenging for more than a single point light source as it requires repeatedly querying the shape MLP for volume density along paths from each scene point to each light source. Moreover, general scenes can be illuminated by light arriving from \emph{all} directions, and addressing this is imperative to recovering relightable representations in unconstrained scenarios. Simulating even simple global illumination in a brute-force manner is intractable: rendering a \emph{single ray} in our scenes under one-bounce indirect illumination with brute-force sampling would require a \emph{petaflop} of computation, and we need to render roughly a \emph{billion} rays over the course of training.

We ameliorate this issue by replacing several brute-force volume density integrals with learned approximations. We introduce a ``visibility'' MLP that emits an approximation of the environment lighting visibility at any input location along any input direction, as well as an approximation of the expected termination depth of the corresponding ray: $\text{MLP}_{\phi}:(\mathbf{x},\boldsymbol{\omega}) \to (\tilde{V}_{\phi}, \tilde{D}_{\phi})$. When rendering, we use these MLP-approximated quantities in place of their actual values:
\begin{align}
\!\!\!\!V(\mathbf{x},\boldsymbol{\omega}) &= \expo{\!-\!\int_{0}^{\infty}\!\!\!\!\!\absrp(\mathbf{x}+s\boldsymbol{\omega})\,ds\!}\,,
\label{eq:truevis}
\\
\!\!\!\!D(\mathbf{x},\boldsymbol{\omega}) &= \!\!\int_{0}^{\infty}\!\!\! \expo{\!-\!\int_{0}^{t}\!\!\!\absrp(\mathbf{x}+s\boldsymbol{\omega})\,ds\!} t \absrp(\mathbf{x} + t \boldsymbol\omega)\,dt \,.
\label{eq:truedepth}
\end{align}
In Section~\ref{subsec:training} we place losses on the visibility MLP outputs $(\tilde{V}_{\phi}, \tilde{D}_{\phi})$ to encourage them to resemble the $(V, D)$ corresponding to the current state of the shape MLP.

Below, we provide a detailed walkthrough of how our Neural Visibility Field approximations simplify the volume rendering integral computation. Figure~\ref{fig:bean} is provided for reference. We first decompose the reflected radiance $L_r(\mathbf{x},\boldsymbol{\omega}_o)$ into its direct and indirect illumination components.
Let us define $L_e(\mathbf{x},\boldsymbol{\omega}_i)$ as radiance due to a light source arriving at point $\mathbf{x}$ from direction $\boldsymbol{\omega}_i$. As defined in Equation~\ref{eqn:reflect_render1}, $L(\mathbf{x},\boldsymbol{\omega}_i)$ is the estimated incoming radiance at location $\mathbf{x}$ from direction $\boldsymbol{\omega}_i$. This means the incident illumination $L_i$ decomposes into $L_e + L$ (direct plus indirect light). The shading calculation for $L_r$ then becomes:
\begin{gather}
\resizebox{0.9\linewidth}{!}{$\displaystyle%
L_r(\mathbf{x},\boldsymbol{\omega}_o)
= \int_{\mathcal{S}}\left(L_e(\mathbf{x},\boldsymbol{\omega}_i)+L(\mathbf{x},-\boldsymbol{\omega}_i)\right)R(\mathbf{x},\boldsymbol{\omega}_i,\boldsymbol{\omega}_o)d\boldsymbol{\omega}_i $} \\
\resizebox{\linewidth}{!}{$\displaystyle%
= \underbrace{\int_{\mathcal{S}}L_e(\mathbf{x},\boldsymbol{\omega}_i)R(\mathbf{x},\boldsymbol{\omega}_i,\boldsymbol{\omega}_o)d\boldsymbol{\omega}_i}_\text{component due to direct lighting} + \underbrace{\int_{\mathcal{S}}L(\mathbf{x},-\boldsymbol{\omega}_i)R(\mathbf{x},\boldsymbol{\omega}_i,\boldsymbol{\omega}_o)d\boldsymbol{\omega}_i}_\text{component due to indirect lighting}\,. \nonumber
$}
\end{gather}
To calculate incident direct lighting $L_e$ we must account for the attenuation of the (known) environment map $E$ due to the volume density along the incident illumination ray $\boldsymbol{\omega}_i$:
\begin{align}
    L_e(\mathbf{x},\boldsymbol{\omega}_i) &= V(\mathbf x, \boldsymbol \omega_i) E(\mathbf x, -\boldsymbol \omega_i)\,.
\end{align}
Instead of evaluating $V$ as another line integral through the volume, we use the visibility MLP's approximation $\tilde V_\phi$. With this, our full calculation for the direct lighting component of camera ray radiance $L(\mathbf{c},\boldsymbol{\omega}_o)$ simplifies to:
\begin{equation}
\resizebox{\linewidth}{!}{$\displaystyle%
\int_{0}^{\infty}\!\!\!\!\!V\big(\mathbf{x}(t),\mathbf{c}\big)\absrp\big(\mathbf{x}(t)\big) \!\!\int_{\mathcal{S}}\!\!\tilde V_\phi\big(\mathbf x(t), \boldsymbol \omega_i\big) E\big(\mathbf x(t), \shortminus\boldsymbol \omega_i\big) R\big(\mathbf{x}(t),\boldsymbol{\omega}_i,\boldsymbol{\omega}_o\big)d\boldsymbol{\omega}_i dt\,.
\label{eqn:nerv_direct}
$}
\end{equation}
By approximating the integrals along rays from each point on the camera ray toward each environment direction when computing the color of a pixel due to direct lighting, we have reduced the complexity of rendering with direct lighting from quadratic in the number of samples per ray to linear.
Next, we focus on the more difficult task of accelerating the computation of rendering with indirect lighting, for which a brute force approach would scale cubically with the number of samples per ray. We make two approximations to reduce this intractable computation. First, we approximate the outermost integral (the accumulated radiance reflected towards the camera at each point along the ray) with a single point evaluation by treating the volume as a hard surface located at the expected termination depth $t'=D(\mathbf{c},-\boldsymbol{\omega}_o)$. Note that we do not use the visibility MLP's approximation of $t'$ here, since we are already sampling $\sigma(\mathbf{x})$ along the camera ray.
This reduces the indirect contribution of $L(\mathbf{c},\boldsymbol{\omega}_o)$ to a spherical integral at a single point $\mathbf{x}(t')$:
\begin{align}
    \int_{\mathcal{S}}L\big(\mathbf{x}(t'),-\boldsymbol{\omega}_i\big)R\big(\mathbf{x}(t'),\boldsymbol{\omega}_i,\boldsymbol{\omega}_o\big)\,d\boldsymbol{\omega}_i \,.
\end{align}
To simplify the recursive evaluation of $L$ inside this integral, we limit the indirect contribution to a single bounce, and once again use the hard surface approximation to avoid computing an integral along a ray for each incoming direction:
\begin{equation}
\resizebox{0.88\linewidth}{!}{$\displaystyle%
L(\mathbf{x}(t'),-\boldsymbol{\omega}_i) \approx 
\int_{\mathcal{S}}L_e(\mathbf{x'}(t''),\boldsymbol{\omega}_i')R(\mathbf{x'}(t''),\boldsymbol{\omega}_i',-\boldsymbol{\omega}_i)d\boldsymbol{\omega}_i'\,,
    $}
\end{equation}
where $t'' = \tilde D_\phi(\mathbf x(t'), \boldsymbol\omega_i)$ is the expected intersection depth along the ray $\mathbf{x'}(t'')=\mathbf{x}(t')+t''\boldsymbol\omega_i$ as approximated by the visibility MLP. Thus the expression for the component of camera ray radiance $L(\mathbf{c},\boldsymbol{\omega}_o)$ due to indirect lighting is:
\begin{equation}
\resizebox{\linewidth}{!}{$\displaystyle%
\int\!\!\!\!\!\int_{\mathcal{S}}\!\!L_e(\mathbf{x'}(t''),\boldsymbol{\omega}_i') R(\mathbf{x'}(t''),\boldsymbol{\omega}_i',\shortminus\boldsymbol{\omega}_i)d\boldsymbol{\omega}_i'R(\mathbf{x}(t'),\boldsymbol{\omega}_i,\boldsymbol{\omega}_o)d\boldsymbol{\omega}_i 
\,,$}
\end{equation}
and fully expanding the direct radiance $L_e(\mathbf{x'}(t''),\boldsymbol{\omega}_i')$ incident at each secondary intersection point gives us:
\begin{equation}
\resizebox{\linewidth}{!}{$\displaystyle%
\int\!\!\!\!\!\int_{\mathcal{S}}\!\! \tilde V_\phi\big(\mathbf{x'}(t''),\boldsymbol{\omega}_i'\big) E\big(\mathbf{x'}(t''),-\boldsymbol{\omega}_i'\big) 
R\big(\mathbf{x'}(t''),\boldsymbol{\omega}_i',\shortminus\boldsymbol{\omega}_i\big)d\boldsymbol{\omega}_i' R\big(\mathbf{x}(t'),\boldsymbol{\omega}_i,\boldsymbol{\omega}_o\big)d\boldsymbol{\omega}_i\,,
\label{eqn:nerv_indirect}
$}
\end{equation}
Finally, we can write out the complete volume rendering equation used by NeRV as the sum of Equations~\ref{eqn:nerv_direct} and \ref{eqn:nerv_indirect}:
\begin{align}
\resizebox{\linewidth}{!}{$\displaystyle%
L(\mathbf{c},\boldsymbol{\omega}_o)\!=\!\! \int_{0}^{\infty}\!\!\!\!\!V\big(\mathbf{x}(t),\mathbf{c}\big)\absrp\big(\mathbf{x}(t)\big) \!\!\int_{\mathcal{S}}\!\!\tilde V_\phi\big(\mathbf x(t), \boldsymbol \omega_i\big) E\big(\mathbf x(t), \shortminus\boldsymbol \omega_i\big) R\big(\mathbf{x}(t),\boldsymbol{\omega}_i,\boldsymbol{\omega}_o\big)d\boldsymbol{\omega}_i dt 
$} \nonumber \\
\resizebox{\linewidth}{!}{$\displaystyle%
+ \int\!\!\!\!\!\int_{\mathcal{S}}\!\! \tilde V_\phi\big(\mathbf{x'}(t''),\boldsymbol{\omega}_i'\big) E\big(\mathbf{x'}(t''),\shortminus\boldsymbol{\omega}_i'\big) 
R\big(\mathbf{x'}(t''),\boldsymbol{\omega}_i',\shortminus\boldsymbol{\omega}_i\big)d\boldsymbol{\omega}_i' R\big(\mathbf{x}(t'),\boldsymbol{\omega}_i,\boldsymbol{\omega}_o\big)d\boldsymbol{\omega}_i\,
$}
\label{eqn:nerv_final}
\end{align}
Figure~\ref{fig:problem} illustrates how the approximations made by NeRV reduce the computational complexity of computing direct and indirect illumination from quadratic and cubic (respectively) to linear.
This enables the simulation of direct illumination from environment lighting and one-bounce indirect illumination within the training loop of optimizing a continuous relightable volumetric scene representation.

\subsection{Rendering}

To render a camera ray $\mathbf{x}(t)=\mathbf{c} - t\boldsymbol{\omega}_o$ passing through a NeRV, we estimate the volume rendering integral in Equation~\ref{eqn:nerv_final} using the following procedure: 
\begin{enumerate}[leftmargin=0cm, itemindent=0.45cm, itemsep=0cm, parsep=0.1cm, label=\textbf{\arabic*})]
    \item We draw 256 stratified samples along the ray and query the shape and reflectance MLPs for the volume densities, surface normals, and BRDF parameters at each point: $\sigma = \text{MLP}_{\theta}(\mathbf{x}(t))$, $\mathbf{n} = \nabla_{\mathbf{x}}\text{MLP}_{\theta}(\mathbf{x}(t))$, $(\mathbf{a}, \gamma) = \text{MLP}_{\psi}(\mathbf{x}(t))$.
    \item We shade each point along the ray with direct illumination by estimating the integral in Equation~\ref{eqn:nerv_direct}. First, we generate $E(\mathbf{x}(t),-\boldsymbol{\omega}_i)$ by sampling the known environment lighting on a $12\times24$ grid of directions $\boldsymbol{\omega}_i$ on the sphere around each point. We then multiply this by the predicted visibility $\tilde{V}_{\phi}(\mathbf{x}(t),\boldsymbol{\omega}_i)$ and microfacet BRDF values $R(\mathbf{x}(t),\boldsymbol{\omega}_i,\boldsymbol{\omega}_o)$ at each sampled $\boldsymbol{\omega}_i$, and integrate this product over the sphere to produce the direct illumination contribution.
    \item We shade each point along the ray with indirect illumination by estimating the integral in Equation~\ref{eqn:nerv_indirect}. First, we compute the expected camera ray termination depth $t'=D(\mathbf{c},-\boldsymbol{\omega}_o)$ using the density samples from Step 1. Next, we sample 128 random directions on the upper hemisphere at $\mathbf{x}(t')$ and query the visibility MLP for the expected termination depths along each of these rays $t'' = \tilde D_\phi(\mathbf x(t'), \boldsymbol\omega_i)$ to compute the secondary surface intersection points $\mathbf{x'}(t'')=\mathbf{x}(t')+t''\boldsymbol\omega_i$. We then shade each of these points with direct illumination by following the procedure in Step 2. This estimates the indirect illumination incident at $\mathbf{x}(t')$, which we then multiply by the microfacet BRDF values $R(\mathbf{x}(t'),\boldsymbol{\omega}_i,\boldsymbol{\omega}_o)$ and integrate over the sphere to produce the indirect illumination contribution.
    \item The total reflected radiance at each point along the camera ray $L_r(\mathbf{x}(t),\boldsymbol{\omega}_o)$ is the sum of the quantities from Steps 2 and 3. We composite these along the ray to compute the pixel color using the same quadrature rule~\cite{max95} used in NeRF: 
\begin{gather}
\resizebox{0.88\linewidth}{!}{$\displaystyle%
L(\mathbf{c},\boldsymbol{\omega}_o)=\sum_{t}V(\mathbf{x}(t),\mathbf{c}) \alpha(\absrp(\mathbf{x}(t)) \delta) L_r(\mathbf{x}(t),\boldsymbol{\omega}_o)\,,
$}
\\
\resizebox{\linewidth}{!}{$%
V(\mathbf{x}(t),\mathbf{c})=\expo{- \sum_{s<t} \absrp(\mathbf{x}(s)) \delta}\,, \quad
\alpha(z) = 1-\expo{-z}\,, \nonumber
$}
\end{gather}
\end{enumerate}
where $\delta$ is the distance between samples along the ray.

\subsection{Training and Implementation Details}
\label{subsec:training}

Instead of directly passing 3D coordinates $\mathbf{x}$ and direction vectors $\boldsymbol{\omega}$ to the MLPs, we map these inputs using NeRF's positional encoding~\cite{mildenhall2020nerf,tancik2020fourfeat}, with a maximum frequency of $2^7$ for 3D coordinates and $2^4$ for 3D direction vectors. The shape and reflectance MLPs each use 8 fully-connected ReLU layers with 256 channels. The visibility MLP uses 8 fully-connected ReLU layers with 256 channels each to map the encoded 3D coordinates $\mathbf{x}$ to an 8-dimensional feature vector which is concatenated with the encoded 3D direction vector $\boldsymbol{\omega}$ and processed by 4 fully-connected ReLU layers with 128 channels each. 

We train a separate NeRV representation from scratch for each scene, which requires a set of posed RGB images and corresponding lighting environments. At each training iteration we randomly sample a batch of 512 pixel rays $\mathcal{R}$ from the input images and use the previously-described procedure to render these pixels from the current NeRV model. We additionally sample 256 random rays $\mathcal{R}'$ per training iteration that intersect the volume, and we compute the visibility and expected termination depth at each location and in either direction along each ray for use as supervision for the visibility MLP. We minimize the sum of three losses:
\begin{gather}
\resizebox{0.5\linewidth}{!}{$\displaystyle%
\mathcal{L}=\sum_{\mathbf{r}\in\mathcal{R}}\norm{\tau(\tilde{L}(\mathbf{r})) - \tau(L(\mathbf{r}))}_2^2 +
$} \quad\quad\quad\quad\quad \\
\resizebox{\linewidth}{!}{$\displaystyle%
\vismult\!\!\!\sum_{\mathbf{r'}\in\mathcal{R'}\cup\mathcal{R}, t}\!\!\left(\norm{\tilde{V}_{\phi}(\mathbf{r'}(t)) -  V_{\theta}(\mathbf{r'}(t))}_2^2 + \norm{\tilde{D}_{\phi}(\mathbf{r'}(t)) - D_{\theta}(\mathbf{r'}(t))}_2^2 \right)\,, \nonumber
\label{eqn:training_loss}
$}
\end{gather}
where $\tau(x)=\nicefrac{x}{1+x}$ is a tone-mapping operator~\cite{gharbi2019sample}, $L(\mathbf{r})$ and $\tilde{L}(\mathbf{r})$ are the ground truth and predicted camera ray radiance values (ground-truth values are simply the colors of input image pixels), $\tilde{V}_{\phi}(\mathbf{r})$ and $\tilde{D}_{\phi}(\mathbf{r})$ are the predicted visibility and expected termination depth from our visibility MLP given its current weights $\phi$,  $V_{\theta}(\mathbf{r})$ and ${D}_{\theta}(\mathbf{r})$ are the estimates of visibility and termination depth implied by the shape MLP given its current weights $\theta$, and $\vismult=20$ is the weight of the loss terms encouraging the visibility MLP to be consistent with the shape MLP. Note that the visibility MLP is not supervised using any ``ground truth'' visibility or termination depth --- it is only optimized to be consistent with the NeRV's current estimate of scene geometry, by evaluating Equations~\ref{eq:truevis} and \ref{eq:truedepth} using the densities $\sigma$ emitted by the shape $\text{MLP}_\theta$. 
We apply a ``stop gradient'' to $V_{\theta}$ and $D_{\theta}$ in the last two terms of the loss, so the shape MLP is not encouraged to degrade its own performance to better match the output from the visibility MLP.
We implement our model in JAX~\cite{jax2018github}, and optimize using Adam~\cite{KingmaB15} with a learning rate that begins at $10^{-5}$ and decays exponentially to $10^{-6}$ over the course of optimization (the other Adam hyperparameters are default values: $\beta_1=0.9$, $\beta_2=0.999$, and $\epsilon=10^{-8}$). Each model is trained for 1 million iterations using 128 TPU cores, which takes $~\!1$ day to converge.

\section{Results}
\label{sec:results}

NeRV outperforms prior work, particularly in its ability to recover relightable scene representations from images observed under complex lighting.
We urge the reader to view our supplementary video to appreciate NeRV's relighting and view synthesis results. 
In Table~\ref{table:results} we show performance for rendering images from novel viewpoints with lighting conditions not observed during training.
We evaluate two versions of NeRV: \textbf{NeRV with Neural Visibility Fields (NeRV, NVF)} and \textbf{NeRV with Test-time Tracing (NeRV, Trace)}. Both methods use the same training procedure as described above, and differ only in how evaluation is performed: ``NeRV, NVF'' uses the same visibility approximations used during training at test time, while ``NeRV, Trace'' uses brute-force tracing to estimate visibility to point light sources to render sharper shadows at test time. We compare against the following baselines:

\newcommand{\grayline}{\arrayrulecolor[rgb]{0.8,0.8,0.8}\hline\arrayrulecolor[rgb]{0,0,0}}
\newcommand{\tablespace}{\,\,\,\,}
\newcommand{\halftablespace}{\,}
\setlength{\tabcolsep}{4pt}
\begin{table}[t]
\centering

\resizebox{\columnwidth}{!}{
\begin{tabular}{l|c|c|c|c|c|c|c|c}
\multicolumn{9}{c}{\textbf{Hotdogs}} \\
Train Illum. & \multicolumn{2}{c|}{Single Point} & \multicolumn{2}{c|}{Colorful + Point} & \multicolumn{2}{c|}{Ambient + Point} & \multicolumn{2}{c}{OLAT} \\
\hline
 & PSNR & MS-SSIM & PSNR & MS-SSIM & PSNR & MS-SSIM & PSNR & MS-SSIM  \\
\hline
NLT~\cite{zhang2020neural} & $-$ & $-$ & $-$ & $-$ & $-$ & $-$ & $23.57$ & $0.851$ \\
NeRF+LE & $19.96$ & $\best{0.868}$ & $17.88$ & $0.758$ & $20.72$ & $0.869$ & $-$ & $-$ \\
NeRF+Env & $19.94$ & $0.863$ & $19.17$ & $0.824$ & $20.56$ & $0.864$& $-$ & $-$  \\
Bi \etal~\cite{bi2020neural} & $23.74$ & $0.862$ & $22.09$ & $0.799$ & $20.94$ & $0.754$ & $-$ & $-$ \\
\grayline
NeRV, NVF & $\best{23.93}$ & $0.860$ & $\best{24.37}$ & $0.885$ & $\best{25.14}$ & $\best{0.892}$ & $-$ & $-$ \\
NeRV, Trace & $23.76$ & $0.863$ & $24.24$ & $\best{0.886}$ & $25.06$ & $\best{0.892}$ & $-$ & $-$ \\
\end{tabular}
}

\vspace{3mm}

\resizebox{\columnwidth}{!}{
\begin{tabular}{l|c|c|c|c|c|c|c|c}
\multicolumn{9}{c}{\textbf{Lego}} \\
Train Illum. & \multicolumn{2}{c|}{Single Point} & \multicolumn{2}{c|}{Colorful + Point} & \multicolumn{2}{c|}{Ambient + Point} & \multicolumn{2}{c}{OLAT} \\
\hline
 & PSNR & MS-SSIM & PSNR & MS-SSIM & PSNR & MS-SSIM & PSNR & MS-SSIM  \\
\hline
NLT~\cite{zhang2020neural} & $-$ & $-$ & $-$ & $-$ & $-$ & $-$ & $24.10$ & $0.936$ \\
NeRF+LE & $21.42$ & $0.874$ & $21.74$ & $0.890$ & $20.33$ & $0.860$ & $-$ & $-$ \\
NeRF+Env & $21.13$ & $0.855$ & $20.27$ & $0.878$ & $20.24$ & $0.852$ & $-$ & $-$ \\
Bi \etal~\cite{bi2020neural} & $22.89$ & $\best{0.897}$ & $22.83$ & $0.890$ & $18.10$ & $0.783$ & $-$ & $-$ \\
\grayline
NeRV, NVF & $22.78$ & $0.866$ & $23.82$ & $0.899$ & $23.32$ & $0.894$ & $-$ & $-$ \\
NeRV, Trace & $\best{23.16}$ & $0.883$ & $\best{24.18}$ & $\best{0.925}$ & $\best{23.79}$ & $\best{0.923}$ & $-$ & $-$ \\
\end{tabular}
}

\vspace{3mm}

\resizebox{\columnwidth}{!}{
\begin{tabular}{l|c|c|c|c|c|c|c|c}
\multicolumn{9}{c}{\textbf{Armadillo}} \\
Train Illum. & \multicolumn{2}{c|}{Single Point} & \multicolumn{2}{c|}{Colorful + Point} & \multicolumn{2}{c|}{Ambient + Point} & \multicolumn{2}{c}{OLAT} \\
\hline
 & PSNR & MS-SSIM & PSNR & MS-SSIM & PSNR & MS-SSIM & PSNR & MS-SSIM \\
\hline
NLT~\cite{zhang2020neural} & $-$ & $-$ & $-$ & $-$ & $-$ & $-$ & $21.62$ & $0.900$ \\
NeRF+LE & $20.35$ & $0.881$ & $18.76$ & $0.863$ & $17.35$ & $0.859$ & $-$ & $-$ \\
NeRF+Env & $19.60$ & $0.874$ & $17.89$ & $0.863$ & $17.28$ & $0.851$ & $-$ & $-$ \\
Bi \etal~\cite{bi2020neural} & $\best{22.35}$ & $0.894$ & $21.06$ & $0.892$ & $19.93$ & $0.842$ & $-$ & $-$ \\
\grayline
NeRV, NVF & $21.14$ & $0.882$ & $22.80$ & $0.910$ & $22.80$ & $\best{0.897}$ & $-$ & $-$ \\
NeRV, Trace & $22.14$ & $\best{0.897}$ & $\best{23.02}$ & $\best{0.921}$ & $\best{22.81}$ & $0.895$ & $-$ & $-$ \\
\end{tabular}
}

\caption{Quantitative relighting and view synthesis results. For every scene, we train each method on three datasets that contain images of the scene under different illumination conditions, and compare the metrics of all variants on the same testing dataset. Please refer to Section~\ref{sec:results} for details.
}
\label{table:results}  
\end{table}
\setlength{\tabcolsep}{1.4pt}

\begin{figure}[t]
\begin{center}
    \includegraphics[width=1\linewidth]{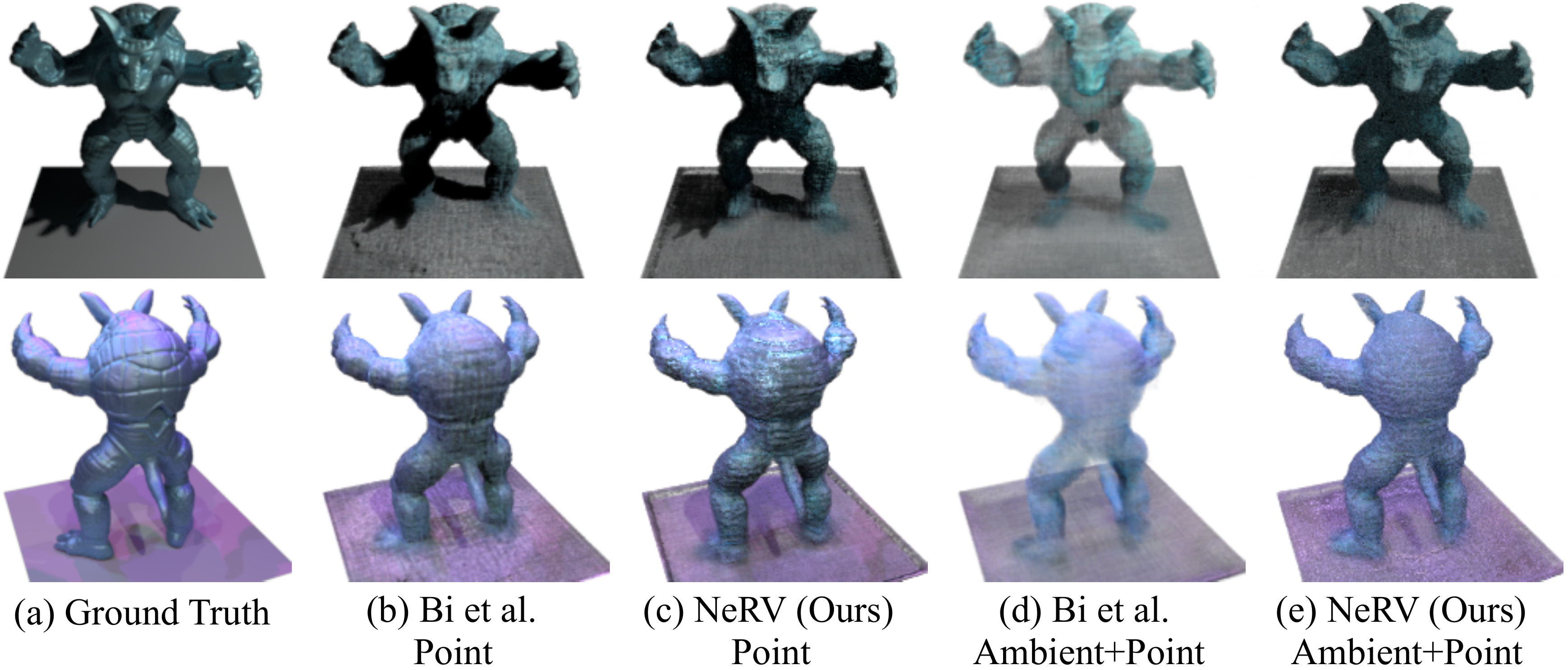}
\end{center}
   \vspace{-0.2in}
   \caption{
   Both (b) Bi \etal~\cite{bi2020neural} and (c) NeRV recover high-quality relightable models when trained on images illuminated by a single point source. However, for more complex lighting such as ``Ambient+Point'', (d) Bi \etal fails as its brute-force visibility computation is unable to simulate the surrounding ambient lighting during training. Their model minimizes training loss by making the scene transparent and is thus unable to render convincing images for the ``single point light'' (row 1) or ``colorful set of points lights'' (row 2) conditions.
   (e) Because NeRV correctly simulates light transport, its renderings more closely resemble (a) the ground truth.
   }
   \label{fig:bi_comparison}
\end{figure}

\begin{figure}[t]
\begin{center}
    \includegraphics[width=1\linewidth]{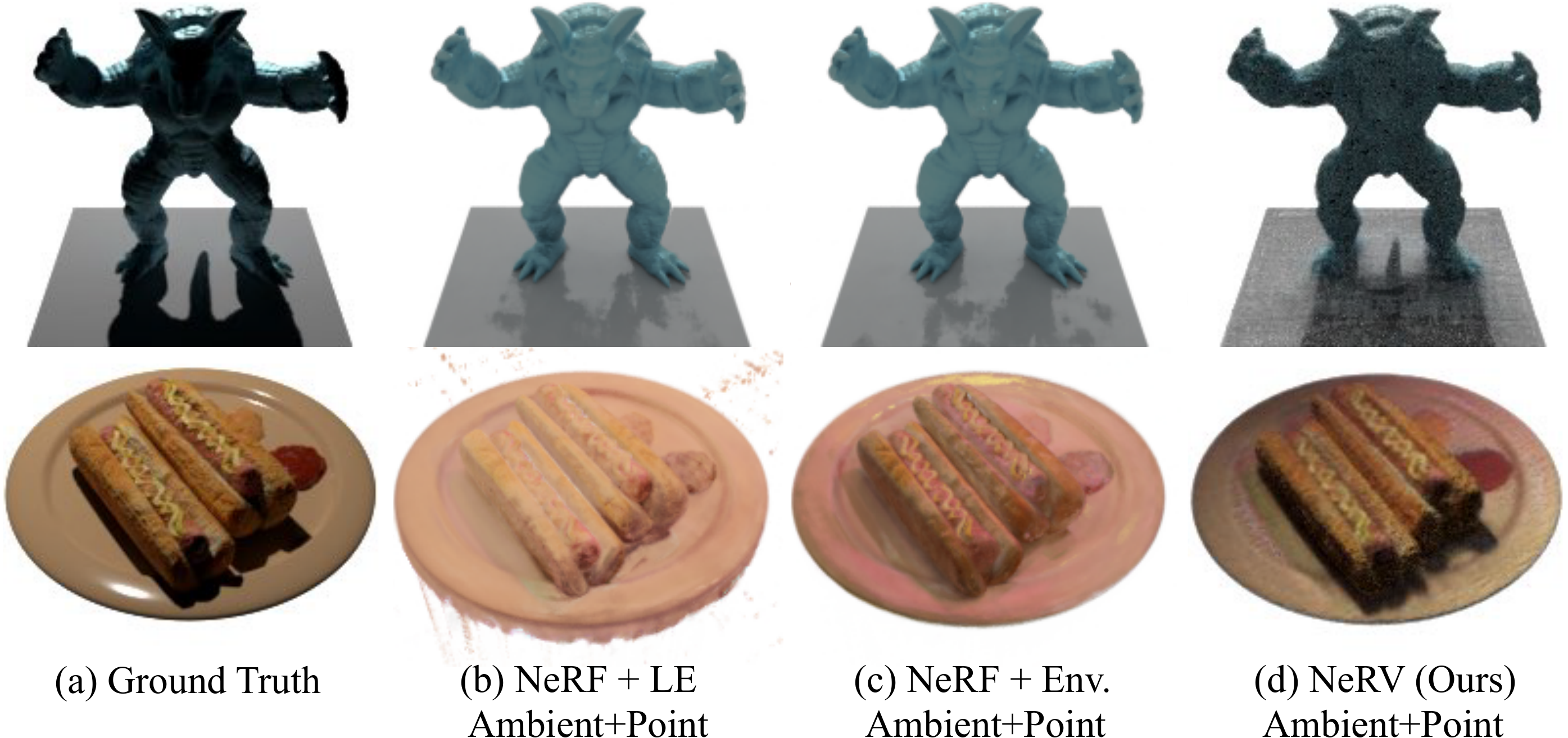}
\end{center}
   \vspace{-0.2in}
   \caption{
   Modeling appearance changes due to lighting with a latent code does not generalize to lighting conditions unlike those seen during training. Here we train (b, c) the two latent code baselines (d) and NeRV on the ``Ambient+Point'' dataset. The latent code models are unable to produce convincing renderings at test time, while NeRV trained on the same data renders high-quality images.
   }
   \label{fig:latent_comparison}
\end{figure}

\textbf{Neural Light Transport~\cite{zhang2020neural} (NLT)} requires an input proxy geometry (which we provide by running marching cubes~\cite{lorensen1987marching} on NeRFs~\cite{mildenhall2020nerf} trained from images of each scene rendered with fixed lighting), and trains a convolutional network defined in an object's texture atlas space to perform simultaneous relighting and view synthesis. Though our method just requires images with known but unconstrained lighting conditions for training, NLT requires multi-view images captured ``One-Light-at-a-Time'' (OLAT), where each viewpoint is rendered multiple times, once per light source. See the supplemental material for qualitative comparisons. 

\textbf{NeRF + Learned Embedding (NeRF+LE)} and \textbf{NeRF + Fixed Environment Embedding (NeRF+Env)} represent appearance variation due to changing lighting using latent variables. Both augment the original NeRF model with an additional input of a 64-dimensional latent code corresponding to the scene lighting condition. These approaches are similar to ``NeRF in the Wild''~\cite{martin2020nerf}, which also uses a latent code to describe appearance variation due to variable lighting. NeRF+LE uses a PointNet~\cite{qi2017pointnet} encoder to embed the position and color of each light, and NeRF+Env simply uses a flattened environment map as the latent code.

\textbf{Neural Reflectance Fields (Bi \etal~\cite{bi2020neural})} uses a similar neural volumetric representation as NeRV, with the critical difference that brute-force raymarching is used to compute visibilities. This approach is therefore unable to consider illumination from sources other than a single point light during training. At test time, when time and memory constraints are less restrictive, it computes visibilities to all light sources.

We train each method (other than NLT) on nine datasets. Each consists of 150 images of a synthetic scene (``Hotdogs'', ``Lego'', or ``Armadillo'') illuminated by one of 3 lighting conditions: 1) ``Point'' contains a single white point light randomly sampled on a hemisphere above the scene for each frame, representing a laboratory setup similar to that of Bi \etal~\cite{bi2020neural}. 2) ``Colorful + Point'' contains a randomly-sampled point light, as well as a set of 8 colorful point lights whose locations and colors are fixed across all images in the dataset. This represents a challenging scenario with multiple strong light sources that cast shadows and tint the scene. 3) ``Ambient + Point'' contains a randomly-sampled point light, as well as a dim grey environment map. This represents a challenging scenario where scene points are illuminated from all directions. We separately train each method on each of these nine datasets and measure performance on the corresponding scene's test set, which consists of 150 images of the scene under novel lighting conditions (containing either one or eight point sources) not observed during training, rendered from novel viewpoints not observed during training.

\subsection{Discussion}

Our method outperforms all baselines in experiments that correspond to challenging complex lighting conditions, and matches the performance of prior work in experiments with simple lighting.
As visualized in Figure~\ref{fig:bi_comparison}, the method of Bi~\etal performs comparably to ours in the case it is designed for: images illuminated by a single point source. However, their model's performance degrades when it is trained on datasets that have complex lighting conditions (``Colorful+Point'' and ``Ambient+Point'' experiments in Table~\ref{table:results}), as its forward model is unable to simulate light from more than a single source during training. As visualized in Figure~\ref{fig:latent_comparison}, our method thoroughly outperforms both latent code baselines, as they are unable to generalize to lighting conditions that are unlike those seen during training. Our method generally matches or outperforms the NLT baseline, which requires a controlled laboratory lighting setup and substantially more inputs than all other methods (the multi-view OLAT dataset we use to train NLT contains eight times as many images as our other datasets, and the original NLT paper~\cite{zhang2020neural} uses 150 OLAT images per viewpoint).

\subsection{Ablation Studies}
\label{sec:ablations}

\begin{figure}[t]
\begin{center}
    \includegraphics[width=1\linewidth]{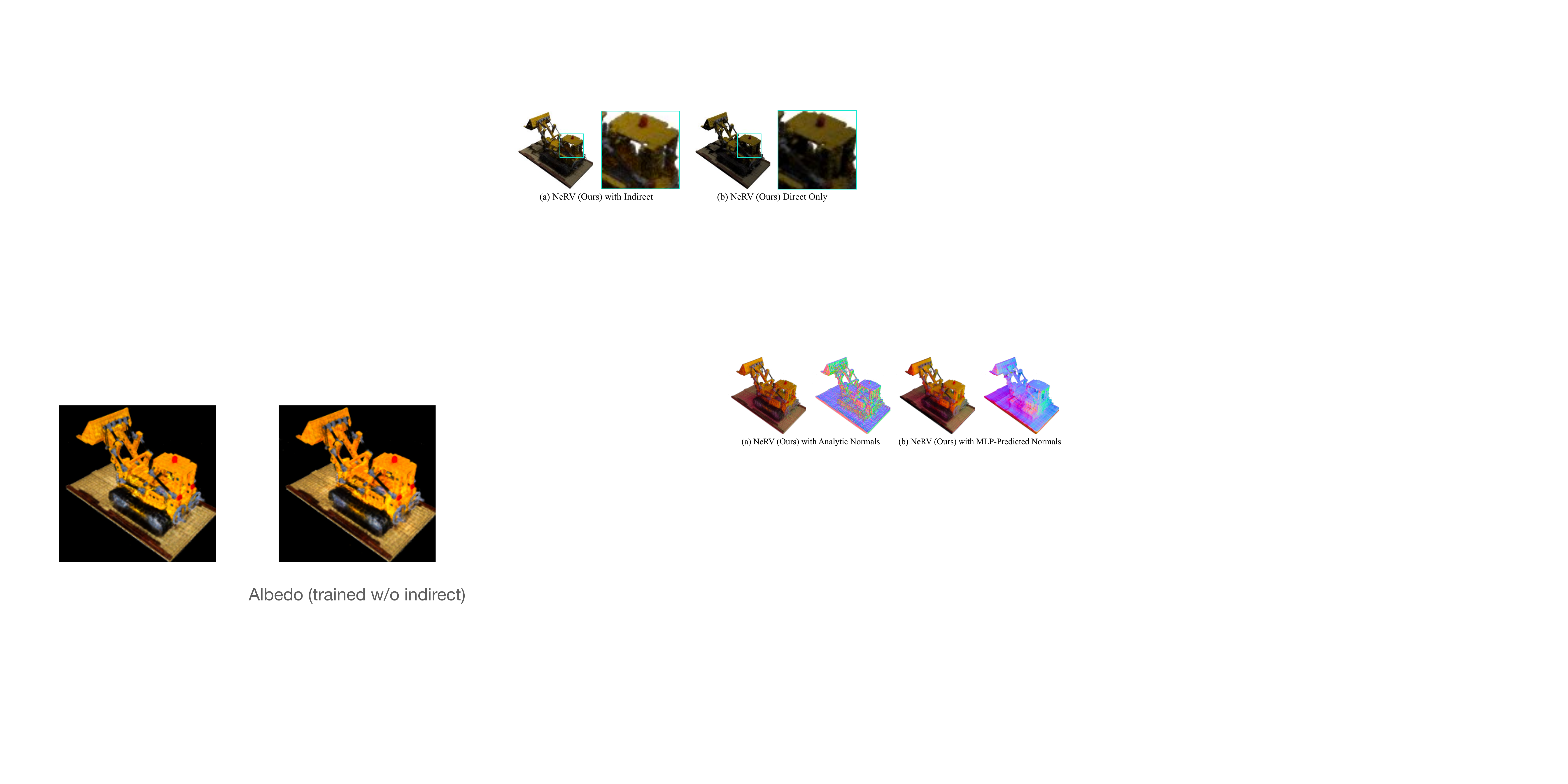}
\end{center}
   \vspace{-0.2in}
   \caption{
   NeRV's ability to simulate indirect illumination produces realistic details such as the additional brightness in the lego bulldozer's cab due to interreflections.
   }
   \label{fig:indirect}
\end{figure}

\begin{figure}[t]
\begin{center}
    \includegraphics[width=1\linewidth]{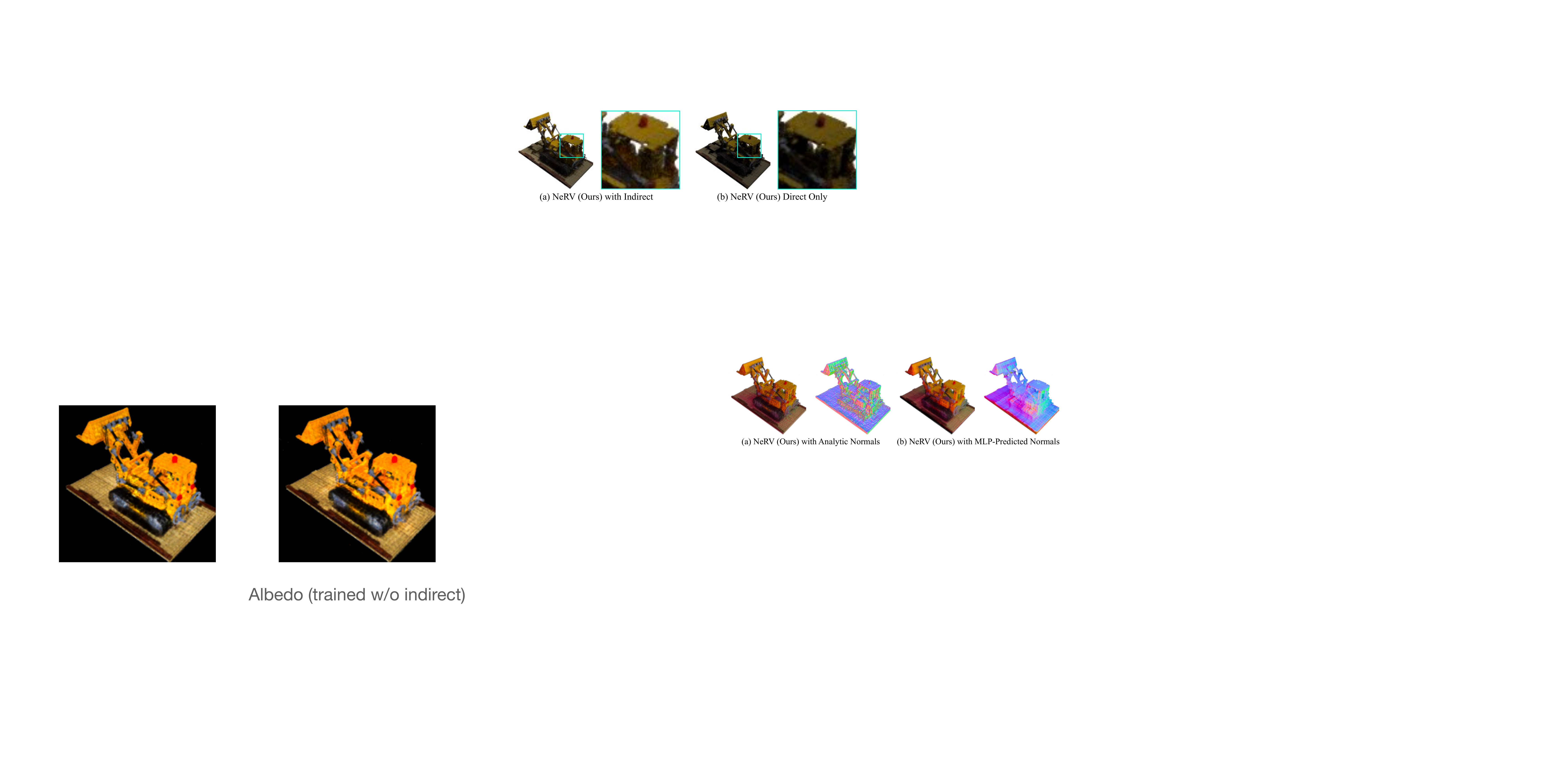}
\end{center}
   \vspace{-0.2in}
   \caption{
   While obtaining surface normals (a) analytically or (b) as an output of the shape MLP produces similar renderings, analytic normals are much closer to the true surface normals.
   }
   \label{fig:normals}
\end{figure}

We validate our choices of using analytic instead of MLP-predicted surface normals and simulating one-bounce indirect illumination through the ablation study reported in Table~\ref{table:ablations}. We can see that modeling indirect illumination improves performance (Figure~\ref{fig:indirect}), even for our relatively simple scenes. Although using analytic instead of MLP-predicted normals is less numerically significant, Figure~\ref{fig:normals} shows that it results in more accurate estimated surface normals, which may be important for downstream graphics tasks.

\setlength{\tabcolsep}{4pt}
\begin{table}[t]
\centering

\resizebox{\columnwidth}{!}{
\begin{tabular}{l|c|c|c|c|c|c}
Scene & \multicolumn{2}{c|}{Hotdogs} & \multicolumn{2}{c|}{Lego} & \multicolumn{2}{c}{Armadillo} \\
\hline
 & PSNR & MS-SSIM & PSNR & MS-SSIM & PSNR & MS-SSIM \\
\hline
Ours, No Indirect & $24.43$ & $0.861$ & $23.06$ & $0.888$ & $21.27$ & $0.878$ \\ 
Ours, MLP Normals & $\best{25.60}$ & $\best{0.893}$ & $23.18$ & $0.886$ & $22.40$ & $0.891$ \\ 
Ours, NVF & $25.14$ & $0.892$ & $23.32$ & $0.894$ & $22.80$ & $\best{0.897}$ \\ 
Ours, Trace & $25.06$ & $0.892$ & $\best{23.79}$ & $\best{0.923}$ & $\best{22.81}$ & $0.895$ \\ 
\end{tabular}
}

\caption{Quantitative ablation results trained on the ``Ambient+Point'' dataset. Please refer to Section~\ref{sec:ablations} for details.
}

\label{table:ablations}  
\end{table}
\setlength{\tabcolsep}{1.4pt}

\section{Conclusion}

We have demonstrated a method for recovering relightable neural volumetric representations from images of scenes illuminated by environmental and indirect lighting, by using a visibility MLP to approximate portions of the volume rendering integral that would otherwise be intractable to estimate during training by brute-force sampling. We believe that this work is an important initial foray into leveraging learned function approximation to alleviate the computational burden incurred by using rigorous physically-based differentiable rendering procedures for inverse rendering.

{\small
\bibliographystyle{ieee_fullname}
\bibliography{egbib}
}

\clearpage

% This version of CVPR template is provided by Ming-Ming Cheng.
% Please leave an issue if you found a bug:
% https://github.com/MCG-NKU/CVPR_Template.

% \documentclass[review]{cvpr}
% %\documentclass[final]{cvpr}

% \usepackage{times}
% \usepackage{epsfig}
% \usepackage{graphicx}
% \usepackage{amsmath}
% \usepackage{amssymb}
% \usepackage{color}
% \usepackage{subcaption}
% \usepackage{mathtools}
% \usepackage{paralist} % compactenum
% \usepackage{enumitem}
% \usepackage{colortbl}
% \usepackage{nicefrac}
% \usepackage{multirow}
% \usepackage[table,dvipsnames]{xcolor}
% \usepackage{booktabs}
% \usepackage{makecell}
% \usepackage{gensymb}
% \usepackage{microtype}
% \usepackage{cancel}
% \usepackage{wasysym}

% \input{macros.tex}

% Include other packages here, before hyperref.

% If you comment hyperref and then uncomment it, you should delete
% egpaper.aux before re-running latex.  (Or just hit 'q' on the first latex
% run, let it finish, and you should be clear).
% \usepackage[pagebackref=true,breaklinks=true,colorlinks,bookmarks=false]{hyperref}

% \usepackage[font=small]{caption}

% \def\cvprPaperID{2454} % *** Enter the CVPR Paper ID here
% \def\confYear{CVPR 2021}
%\setcounter{page}{4321} % For final version only

% \begin{document}

%%%%%%%%% TITLE
\appendix
% \title{NeRV: Neural Reflectance and Visibility Fields for Relighting and View Synthesis \\ Supplementary Materials}

% \and

% \maketitle

This document contains additional implementation details for our method, as well as additional qualitative results from the experiments discussed in the main paper. Please view our included supplementary video for a brief overview of our method, qualitative results with smoothly-moving novel light and camera paths, and demonstrations of additional graphics applications.

\section{BRDF Parameterization}

We use the standard microfacet bi-directional reflectance distribution function (BRDF) described by Walter \etal~\cite{walter2007microfacet} as our reflectance function, and incorporate some of the simplifications discussed in the BRDF implementations of the Filament~\cite{filamentbrdf} and Unreal Engine~\cite{unrealbrdf} renderering engines. The BRDF $R(\mathbf{x}, \boldsymbol{\omega}_i, \boldsymbol{\omega}_o)$ we use is defined for any 3D location $\mathbf{x}$, incoming lighting direction $\boldsymbol{\omega}_i$, and outgoing reflection direction $\boldsymbol{\omega}_o$ as:
\begin{align}
R(\mathbf{x}, \boldsymbol{\omega}_i, \boldsymbol{\omega}_o)=&\frac{D(\mathbf{h}, \mathbf{n}, \gamma)F(\boldsymbol{\omega}_i, \mathbf{h})G(\boldsymbol{\omega}_i, \boldsymbol{\omega}_o, \gamma)}{4(\mathbf{n} \cdot \boldsymbol{\omega}_o)} \\ & +(\mathbf{n} \cdot \boldsymbol{\omega}_i)(1-F(\boldsymbol{\omega}_i, \mathbf{h}))\left(\frac{\mathbf{a}}{\pi}\right)\,, \nonumber \\
D(\mathbf{h}, \mathbf{n}, \gamma)=&\frac{\rho^2}{\pi((\mathbf{n}\cdot\mathbf{h})^2(\rho^2-1)+1)^2}\,, \\
F(\boldsymbol{\omega}_i, \mathbf{h})=&F_0 + (1-F_0)(1-(\boldsymbol{\omega}_i \cdot \mathbf{h}))^5\,, \\
G(\boldsymbol{\omega}_i, \boldsymbol{\omega}_o, \gamma)=&\frac{(\mathbf{n} \cdot \boldsymbol{\omega}_o)(\mathbf{n} \cdot \boldsymbol{\omega}_i)}{(\!(\mathbf{n} \cdot \boldsymbol{\omega}_o)(1\!-\!k)\!+\!k)(\!(\mathbf{n} \cdot \boldsymbol{\omega}_i)(1\!-\!k)\!+\!k)}, \\
\rho=\gamma^2\,, & \quad\quad \mathbf{h}=\frac{\boldsymbol{\omega}_o + \boldsymbol{\omega}_i}{\norm{\boldsymbol{\omega}_o + \boldsymbol{\omega}_i}}\,, \quad\quad k=\frac{\gamma^4}{2}\,,
\end{align}
where $\mathbf{a}$ is the diffuse albedo, $\gamma$ is the roughness, and $\mathbf{n}$ is the surface normal at 3D point $\mathbf{x}$. We use $F_0=0.04$, which is the typical value of dielectric (non-conducting) materials. Note that our definition of the BRDF includes the multiplication by the Lambert cosine term $(\mathbf{n} \cdot \boldsymbol{\omega}_i)$ in order to simplify the equations in the main paper.

\vfill\eject

\section{Additional Qualitative Results}

\begin{figure}[t]
\begin{center}
    \includegraphics[width=1\linewidth]{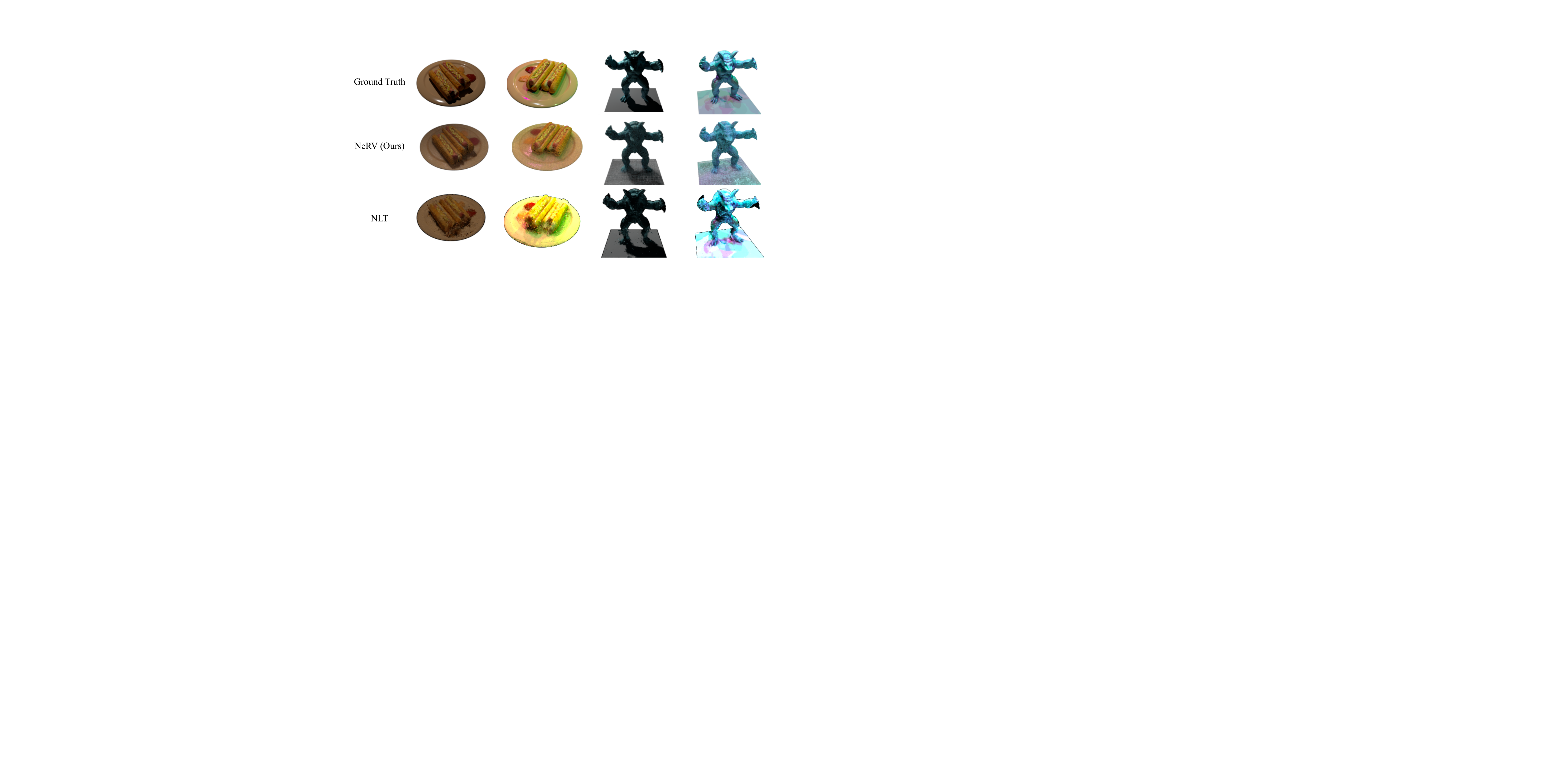}
\end{center}
   \vspace{-0.2in}
   \caption{
   Additional qualitative results, specifically comparing images rendered by NeRV to those rendered by the Neural Light Transport~\cite{zhang2020neural} (NLT) baseline. Note that NLT uses a controlled laboratory lighting setup with eight times as many images as used by NeRV, and an input proxy geometry (which is recovered by training a NeRF~\cite{mildenhall2020nerf} model on a set of images with fixed illumination). The artifacts seen in the shadows of NLT's renderings demonstrate the difference between recovering geometry that works well for view synthesis (as NLT does) and recovering geometry that works well for both view synthesis and relighting (as NeRV does).
   }
   \label{fig:nlt_comparison}
\end{figure}

\begin{figure*}[t]
\begin{center}
    \includegraphics[width=1\linewidth]{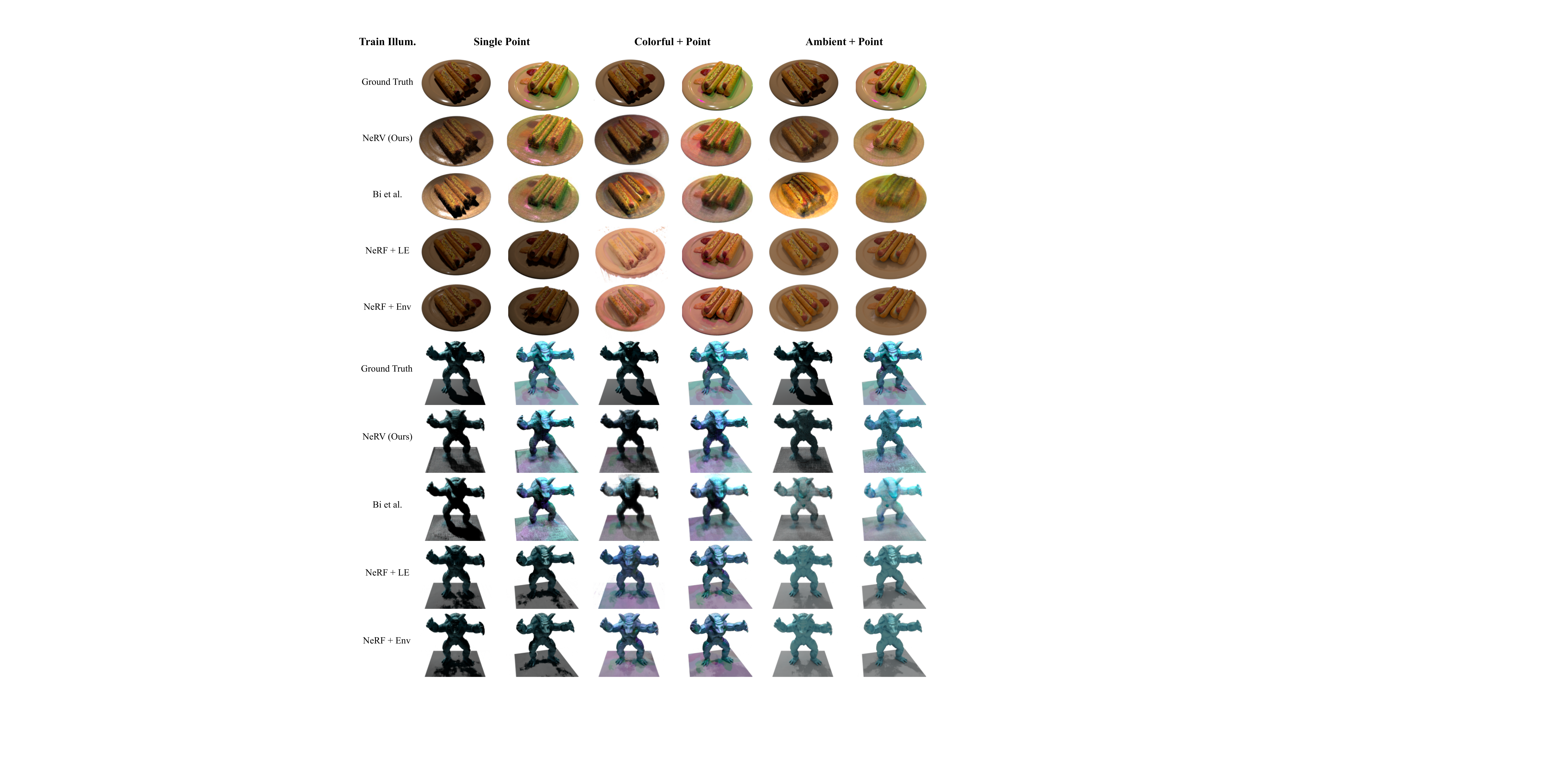}
\end{center}
   \vspace{-0.2in}
   \caption{
   Additional qualitative results from the experiments discussed in the main paper. We can see that NeRV is able to render convincing images from novel viewpoints under novel lighting conditions. The method of Bi \etal~\cite{bi2020neural} is unable to recover accurate models when trained with illumination more complex than a single point light (columns 3-6). Methods that use latent codes to explain variation in appearance due to lighting (NeRF+LE, NeRF+Env) are unable to generalize to lighting conditions different than those seen during training.
   }
   \label{fig:supp_results}
\end{figure*}

Figure~\ref{fig:supp_results} shows additional renderings from NeRV and other baseline methods. We see that NeRV is able to recover effective relightable 3D scene representations from images of scenes with complex illumination conditions. Prior work such as Bi \etal~\cite{bi2020neural} are unable to recover accurate representations from images with lighting conditions more complex than a single point light. Latent code methods (representative of ``NeRF in the Wild''~\cite{martin2020nerf}) are unable to generalize to simulate lighting conditions unlike those seen during training. 

\end{document}